\documentclass[10pt,twocolumn,letterpaper]{article}

\usepackage[pagenumbers]{cvpr} 

\usepackage{float}

\definecolor{cvprblue}{rgb}{0.21,0.49,0.74}
\usepackage[pagebackref,breaklinks,colorlinks,allcolors=cvprblue]{hyperref}

\def\paperID{*****} 
\def\confName{CVPR}
\def\confYear{2025}

\title{\itsname: Bijective Maximum Likelihood Learning
Approach to \\ Hallucination Prediction and Mitigation in Large Vision-Language Models}

\author{Huu-Thien Tran, Thanh-Dat Truong, Khoa Luu\\
CVIU Lab, University of Arkansas\\
{\tt\small \{ht035, tt032, khoaluu\}@uark.edu} \\{\tt\small \url{https://uark-cviu.github.io} }
}

\begin{document}
\maketitle
\begin{abstract}
Large vision-language models have become widely adopted to advance in various domains. However, developing a trustworthy system with minimal interpretable characteristics of large-scale models presents a significant challenge. One of the most prevalent terms associated with the fallacy functions caused by these systems is hallucination, where the language model generates a response that does not correspond to the visual content. To mitigate this problem, several approaches have been developed, and one prominent direction is to ameliorate the decoding process. 
In this paper, we propose a new Bijective Maximum Likelihood Learning (\itsname) approach to hallucination mitigation using normalizing flow theories. The proposed \itsname method can efficiently mitigate the hallucination problem in prevailing vision-language models, resulting in significant improvements. 
Notably, \itsname achieves the average F1 score of $85.06\%$ on POPE benchmark and remarkably reduce $\operatorname{CHAIR}_S$ and $\operatorname{CHAIR}_I$ by $7.6\%$ and $2.6\%$, respectively. 
To the best of our knowledge, this is one of the first studies that contemplates the bijection means to reduce hallucination induced by large vision-language models.
\end{abstract}
\vspace{-2mm}
    
\section{Introduction}
\label{sec:intro}

\begin{figure}
    \centering
    \includegraphics[width=1.0\linewidth]{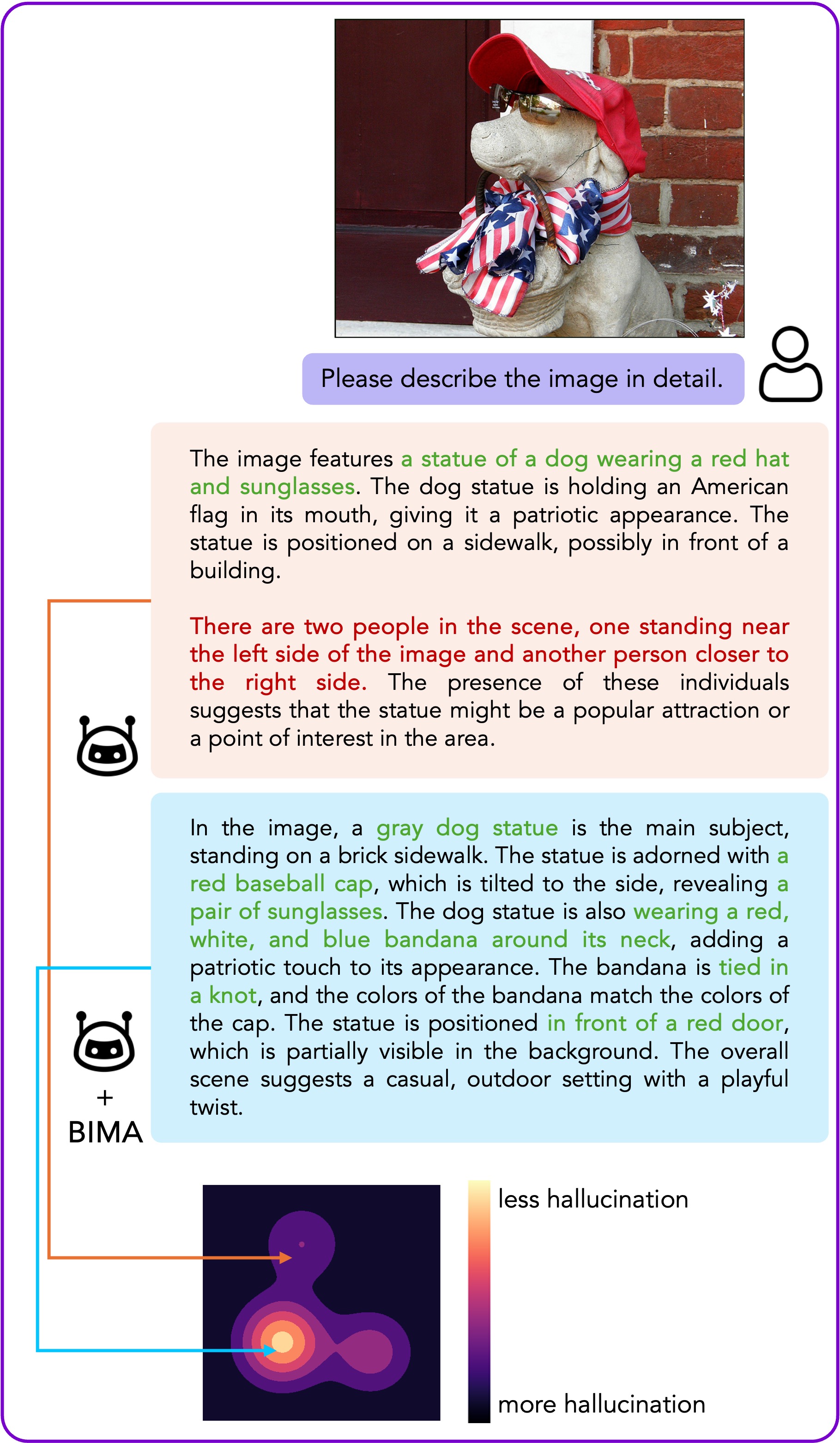}
    \vspace{-4mm}
    \caption{Our \itsname approach helps mitigate hallucinations in large vision-language models. It presents a bijective mapping between each response and its correlative degree of hallucination.}
    \label{fig:teaser}
    \vspace{-8mm}
\end{figure}

Foundation models~\cite{li2024multimodal} have shown their well-rounded prowess in various aspects at an unparalleled level, driven by advancements in modern technology and computational infrastructure. The emergence of large language models~\cite{vicuna2023, touvron2023llama, touvron2023llama2, grattafiori2024llama3, team2024gemma, team2024gemma2, yang2024qwen2, brown2020gpt3} and its impressive capability across various domains has fulfilled the potential to explore the invariability across modalities (\eg visual or auditory data).
Large Vision-Language Models (LVLMs)~\cite{liu2023llava, liu2024improved, dai2023instructblip, zhu2024minigpt4, achiam2023gpt, team2024gemini, Qwen2-VL, Qwen2.5-VL, claude3, guo2025deepseek} have surfaced as a pivotal research area at the intersection of computer vision and natural language processing. 
Users can engage in interactive textual communication with these models through text prompts and visual inputs, either images or video clips. The model will generate contextually relevant and meaningful sentences responding to these prompts. Advancements in model architecture, a broader training dataset, and robust hardware infrastructure have led to continuous improvements in the performance of LVLMs, enabling a broader spectrum of applications.

Despite their phenomenal dexterity, LVLMs suffer from a terminal hindrance known as ``hallucination."
Explicitly, hallucination is the phenomenon where the responses of the LVLM are semantically coherent yet irrelevant and misaligned with the provided visual content and textual prompt as the context~\cite{liu2024survey, sahoo2024survey}. 
This aberration is a considerable concern for reliability when employing these models under critical scenarios, with the risk of spurious decision-making or misleading information. 
For example, in high-stakes healthcare or autonomous driving, such indiscretions are inadmissible and can lead to catastrophic consequences.

Object hallucination~\cite{rohrbach-etal-2018-object-chair, li-etal-2023-evaluating-pope, gunjal2024detecting_mhal} is one of the most extensively studied challenges, where LVLMs generate inconsistent sequences of outputs given the ground-truth objects of the visual component. 
Numerous approaches have been developed to address this fundamental issue~\cite{liu2024survey, sahoo2024survey}. 
Among these approaches, particular attention is given to enhancements in the decoding process~\cite{leng2024vcd, icd, liu2024paying, huo2025sid, suo2025octopus}, offering an efficient means of mitigating the hallucination problem. 
For instance, contrastive decoding leverages the antithetic likelihood measurement at each next-token prediction step, resulting in a remedial mechanism. 
Practically, these contrastive decoding methods introduce uncertainty to the context, either under visual input or textual prompt, to suppress the occurrence of hallucinatory tokens.

To further investigate the impact of the decoding process on the robustness against hallucination, we devise a novel technique that leverages a procedure of learning the underlying distribution of the ground-truth token sequences via bijective maximum likelihood learning. 
We advocate for the creation of a reference distribution that can effectively mitigate the hallucinations induced by LVLMs. 
Consequently, we establish a bijective correspondence between arbitrary model-generated and desirable reference responses, thereby developing an instrument to facilitate decoding. 
We use a flow-based generative model as a robust density distribution estimator. 
This bijective transformation enables a more structured decoding process, seamlessly integrating into the fine-tuning stage of visual instruction-tuned models~\cite{liu2023llava, liu2024improved}.
As a result, our approach helps address the hallucination issue in LVLMs, as illustrated in \cref{fig:teaser}.

Through extensive experiments, our proposed technique demonstrates its effectiveness in reducing hallucinations, thereby improving the reliability of LVLMs across various benchmarks. 
Unlike other decoding strategies, our method requires a fine-tuning process before being adaptively adopted in LVLMs. 
Nevertheless, it offers substantial benefits by being the first to introduce a bijective mapping framework for alleviating hallucinations. 

\noindent
\textbf{Contributions of this Work:} 
In this paper, we propose a novel bijective maximum likelihood learning approach to hallucination mitigation in large vision-language models. Our contributions can be summarized as follows.
First, we establish a comprehensive study advocating a reference distribution of non-hallucinated responses.
Second, we propose a novel decoding approach inspired by bijective mapping using normalizing flow theory based on a novel bijective metric between the reference and model-generated distribution.
Third, we introduce a method for instruction fine-tuning LVLMs to mitigate hallucinations.
Finally, we conduct extensive empirical evaluations to demonstrate the superiority of our proposed \itsname in reducing hallucinations and enhancing the model's trustworthiness.

\section{Related Work}
\label{sec:related}

\subsection{Hallucination Mitigator in LVLMs}
Recent studies have witnessed the emergence of large-scale foundation models equipped to learn across various modalities. 
Since CLIP~\cite{radford2021learning_clip} and BLIP~\cite{li2022blip} assert the ability to effectively align visual and textual representations, numerous subsequent works extend the capability of understanding multi-modality data. 
Thanks to open-sourcing large language models such as LLaMA~\cite{touvron2023llama, touvron2023llama2, grattafiori2024llama3} and Vicuna~\cite{vicuna2023}, several prominent research initiatives have emerged to integrate a visual encoder and a large language model to enhance the learning of alignment across modalities. 
As a result, LVLMs, such as LLaVA~\cite{liu2023llava, liu2024improved}, InstructBLIP~\cite{dai2023instructblip}, MiniGPT-4~\cite{zhu2024minigpt4}, have made significant strides to the field, offering users enhanced interactions with images and texts as prompts. During this phase, LVLMs typically adhere to the same training procedure, compromising two consecutive phases, \ie, pre-training for feature alignment and instruction fine-tuning. The former phase facilitates the alignment of cross-modality representations, while the latter aids in teaching the model to engage in conversations with users. Nevertheless, these LVLMs suffer from a notable hallucination problem, which significantly impacts the reliability of their outputs.

Studies have been conducted on approaches to addressing hallucinations, such as outlined in \cite{liu2024survey, sahoo2024survey}. This paper explores and compares different decoding strategies to deal with hallucinations since this approach has been well-studied in large language models \cite{li-etal-2023-contrastive-text, shi2024trusting_contrast_text}. One notable advantage of this approach is that it serves as the superseded or complementary method to common decoding methods currently employed in language models, including beam search decoding~\cite{NIPS2014_beamsearch, graves2012sequence_beam, freitag-al-onaizan-2017-beam}, Top-$k$ sampling~\cite{angela2018topk}, and Top-$p$ (nucleus) sampling~\cite{Holtzman2020nucleus_sampling}.
Liu~\etal~\cite{liu2024paying} inspected to assess the model's performance under two distinct scenarios: with and without visual element in the input context. This exploration revealed a phenomenon known as ``text inertia,'' which arises from the over-reliance on the language model's context. To address this issue, they designed an algorithm to counteract it. By dynamically adjusting the attention weights assigned to visual tokens, they achieved an equilibrium, giving greater prominence to the imagery representations.
Huang~\etal~\cite{huang2024opera} observed that knowledge aggregation patterns manifest in the self-attention matrix, which is closely associated with the majority of hallucinated tokens in the model’s responses.
In response to this observation, they proposed OPERA, a novel decoding method combining beam search decoding and a penalty term to alleviate the over-trust problem. Additionally, OPERA incorporates a retrospection-allocation strategy, which involves rolling back to previously generated tokens to re-select candidate tokens, thereby preventing the recurrence of such patterns.
Leng~\etal~\cite{leng2024vcd} introduced a simple and training-free approach, called visual contrastive decoding (VCD), that investigates the multimodal alignment by leveraging visual uncertainty. This method creates corresponding distorted visual inputs through a diffusion process, which effectively mitigates the reliance of models on statistical biases and unimodal priors.
Wang~\etal~\cite{icd} presented a different constrastive decoding strategy, dubbed instruction contrastive decoding (ICD). In ICD, the model is perturbed by disturbance instructions. This perturbation allows the model to contrast distributions from standard instructions with the noisy ones, thereby introducing a novel alignment uncertainty to LVLMs.
Huo~\etal~\cite{huo2025sid} revealed the inherent aptitude to assess the significance of visual tokens based on the given context and introduced a method known as self-introspective decoding (SID).
Suo~\etal~\cite{suo2025octopus} recently revisited the approach of employing contrastive decoding by developing a flexible strategy that adaptively identifies hallucination types and constructs a flexible contrastive decoding workflow akin to an octopus's tentacles.
Far apart from previous approaches, Jiang~\etal ~\cite{jiang2025interpreting} applied the logit lens technique~\cite{logitslen} to project the hidden representations of the LVLMs through the lens of their language vocabulary. This enabled them to investigate the causes of hallucinated objects, which were subsequently removed by their introduced knowledge erasure algorithm known as ProjectAway.

\subsection{Flow-based Models}

A flow-based generative model explicitly models a probability distribution harnessing normalizing flows~\cite{rezende2015variational}. 
It is established by a sequence of invertible transformations and utilizes the change-of-variable law to transform a simple distribution into a complex one.
Dinh~\etal~\cite{dinh2017realnvp} proposed a tractable model using real-valued non-volume preserving (RealNVP) transformations that are stably invertible, culminating in an unsupervised learning algorithm of the probabilistic model with exact log-likelihood computation and explainable latent space.
Kingma~\etal~\cite{kingma2018glow} introduced a reversible $1\times 1$ convolution block and a flow-based generative model called Glow to improve log-likelihood and efficiently synthesize high-resolution natural images.
Germain~\etal~\cite{germain2015made} made a minor adjustment to autoencoder neural networks by masking its parameters. It can stimulate the auto-regressive constraint, where each input is reconstituted only from prior inputs in a given sequence.
Kingma~\etal~\cite{kingma2016iaf} proposed inverse auto-regressive flow (IAF), which contains a sequence of invertible transformations, each based on an auto-regressive neural network.
Papamakarios~\etal~\cite{papamakarios2017maf} stacked auto-regressive models, each of which models the successive model's random numbers to obtain a normalizing flow.
This density estimator is called Masked Auto-regressive Flow (MAF), a general form of RealNVP and closely pertains to IAF.
Duong~\etal~\cite{duong2020vec2face} presented a bijective metric learning paradigm for synthesizing the identity of human facials of a given person's features.
Truong~\etal~\cite{truong2021bimal} utilized flow-based philosophy and introduced a bijective maximum likelihood for unsupervised domain adaptation.

\section{The Proposed \itsname Approach}
\label{sec:method}

\begin{figure*}[t]
    \centering
    \includegraphics[width=1\linewidth]{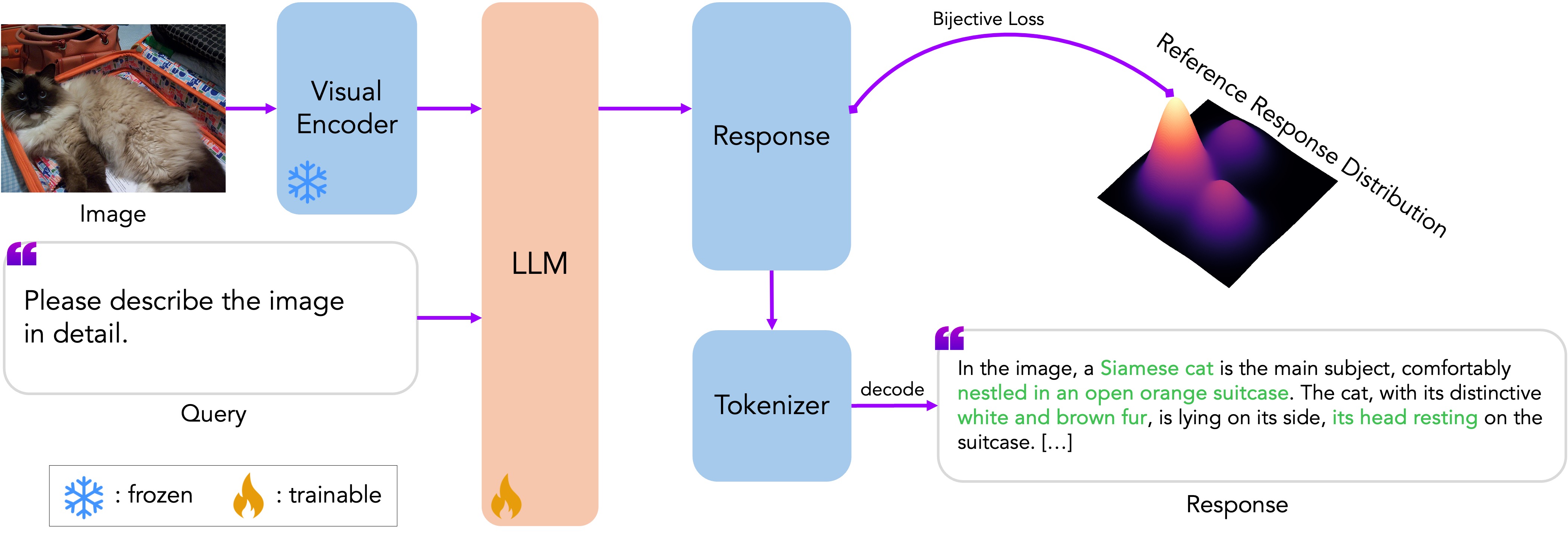}
    \caption{The overview framework of our proposed \itsname, a bijection-inspired decoding approach. The LVLM has two main components: the visual encoder, frozen during training, and the large language model that will be fine-tuned in our approach. In the decoding phase of our \itsname approach, the response will refer to its reference distribution via calculating the bijective loss. Subsequently, the model response can become more accurate before being decoded by the tokenizer.}
    \label{fig:framework}
    \vspace{-4mm}
\end{figure*}

In this section, we first provide an overview of the decoding process in vision-language models and flow-based generative models in \cref{sec:method:preli}. Then, we formalize our proposed \itsname, a bijection-driven model, to mitigate hallucination in LVLMs in \cref{sec:method:flow}. This section compromises our developments with respect to the reference response distribution, how to parameterize it, and a bijective metric for instruction fine-tuning. Finally, we discuss the formation of such reference distribution and the process of fine-tuning with the incorporation of the bijective metric in \cref{sec:method:finetune}.

\subsection{Preliminaries}
\label{sec:method:preli}

{\bfseries Decoding in Vision-Language Models.}
Let $\theta$ be the parameterized LVLM. This model takes the text query tokens $\mathbf x$ and the visual tokens $\mathbf v$ as inputs. Here, we omitted the appearance of system prompt tokens for simplicity. Given a set of queries, the model will produce a response $\mathbf y$, which is generated following an auto-regressive manner under a probabilistic distribution conditioned on $\mathbf x$ and $\mathbf v$. Formally, this process is written as in \cref{eq:eq1}.
\begin{equation} \label{eq:eq1}
    p_\theta(\mathbf y|\mathbf x, \mathbf v) = \prod _{i=0}^{L-1}p_\theta(y_i|\mathbf x, \mathbf v,y_{<i}),
\end{equation}
where $y_i$ represents the token generated at the $i^\text{th}$ time step, $y_{<i}=\{ y_j\}_{j=0}^{i-1}$ is the sequence of preceding tokens up to the time step $(i-1)$, and $L$ is the length of the token sequence of the target $\mathbf y$.

Upon obtaining the logits (or probabilities) $p_\theta(\mathbf y|\mathbf x, \mathbf v)$, several decoding strategies could be employed. 
However, these three methods continue to prevail among all the widely used auto-regressive LVLMs: beam search, top-$k$ sampling, and top-$p$ sampling.
Beam Search~\cite{NIPS2014_beamsearch, graves2012sequence_beam, freitag-al-onaizan-2017-beam} is a score-based algorithm that maintains a limited number of candidates with the highest accumulated scores in a beam.
Top-$k$ Sampling~\cite{angela2018topk} generates the following tokens by randomly selecting from $k$ candidates with the highest likelihood.
Top-$p$ (Nucleus) Sampling~\cite{Holtzman2020nucleus_sampling} enhances Top-$k$ sampling by dynamically considering an amassed number of tokens that hits the probability $p$. 
The decoded token is subsequently appended to the end of the token sequence for the next cycle of generation until the stop criterion is met.

Ultimately, the desirable model $\theta^*$ can be attained by learning to minimize the negative log-likelihood, formulated as in \cref{eq:llm:nll}. 
\small
\begin{equation}
\begin{split}
    \theta^* &=\arg\min_\theta \mathbb E_{(\mathbf x, \mathbf v, \mathbf y) \in \mathcal D} \left[-\log p_\theta(\mathbf y|\mathbf x, \mathbf v) \right]
    \\ &=\arg\min_\theta \mathbb E_{(\mathbf x, \mathbf v, \mathbf y) \in \mathcal D} \left[-\sum_{i=0}^{L-1} \log \left(p_\theta(y_i|\mathbf x, \mathbf v,y_{<i})\right) \right]
\end{split}
\label{eq:llm:nll}
\end{equation}\normalsize
where $\mathcal D$ denotes the dataset used for training the model, which fundamentally consists of pairs of query and answer and the corresponding visual content.

\noindent
{\bfseries Flow-based Generative Models.} 
The fundamental principle for density estimation employing normalizing flows~\cite{rezende2015variational} is to construct an invertible bijection $f:\mathbb R^d \to\mathbb R^d$, \ie there exists an inverse function $g$ such that $g \circ f(z) = z$, in which $\circ$ denotes the function composition operation.
For $i\in\{1,2,\cdots,K\}$, define a sequence of transformed random variables as $z_i=f_i(z_{i-1})$, where the initial variable $z_0$ is distributed according to a known base distribution $p_0(z_0)$.
Each function $f_i$ is required to be invertible to ensure that the transformation remains bijective.
The final variable $z_K$ follows the target distribution $p_K(z_K)$, which is obtained by sequentially applying bijective transformations.
Formally:
\begin{equation}
    z_K=f_K\circ \cdots\circ f_2 \circ f_1 (z_0).
\end{equation}
This constitution facilitates explicit density computation via the change of variables formula. This is because the probability density function of $z_K$ can be expressed in terms of $p_0$ and the Jacobian determinants of the transformations.

\subsection{Our Proposed \itsname Method}
\label{sec:method:flow}

\begin{figure*}
    \centering
    \includegraphics[width=\linewidth]{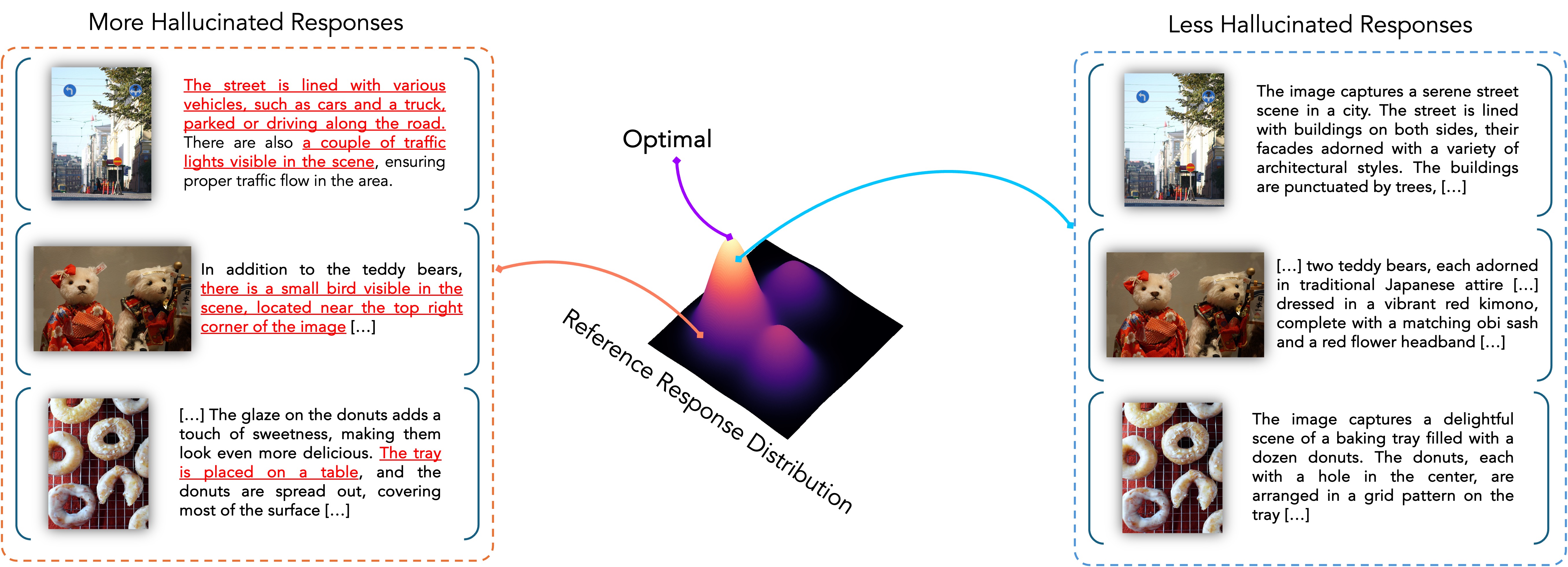}
    
    \caption{Visualization of the Reference Response Distribution $\pi_\mathit{ref}$ in our proposed \itsname. Optimal region of $\pi_\mathit{ref}$ is constructed by ground-truth responses. Any response corresponds to $\pi_\mathit{ref}$ with a degree of hallucination via a bijective mapping function.
    }
    \label{fig:bijec}
    \vspace{-2mm}
\end{figure*}

\begin{remark}[Non-hallucinated Responses]   
\label{rm:nonhallu}
Given a model $\mathcal F$ and a training dataset $\mathcal D$, $\mathcal F$ is an auto-regressive model, and $\mathcal D$ consists of triplets: a query $\mathbf x$, an image $\mathbf v$, and a ground-truth response $\mathbf y$. Thus, predicted responses $\hat{\mathbf y} \approx \mathbf y$ do not induce hallucinations.
\end{remark} 

In \Cref{rm:nonhallu}, we deploy an argument that any responses derived from the given ground-truth dataset do not induce hallucinations during the inference process of the LVLMs. 
These reference responses are implicitly reliable because the model is explicitly instructed to produce token sequences that are analogous to them. 
%
Hallucinatory responses, by contrast, should not follow the same distribution; otherwise, the model would be guided to produce erroneous outputs, undermining its credibility.
%
To formalize this distinction, we study the distribution of non-hallucinated responses, establishing a theoretical foundation for our proposed method.

\begin{remark}[Reference Response Distribution]
\label{rm:ref-dist}
Inherited from \Cref{rm:nonhallu}, a distribution exists where all ground-truth responses $\mathbf{y}$ follow.
This distribution is denoted as $\pi_\mathit{ref}$, which represents the reference response distribution for the ground-truth sequence $\mathbf{y} \in \mathcal{D}$ as in \cref{eq:ref-dist}.
\begin{equation}
\label{eq:ref-dist}
\mathbf y \sim \pi_\mathit{ref}(\mathbf y), \forall \mathbf y \in \mathcal D.
\end{equation} 
\end{remark}

Given the definition of the reference response distribution $\pi_\mathit{ref}$ as outlined in \Cref{rm:ref-dist}, it can be approximated by optimizing the posterior distribution. 
It is noteworthy that since $\mathbf y$ is the variable conditioned on $\mathbf{x}$ and $\mathbf{v}$, the reference response distribution can be fully expressed as $\pi_{\mathit{ref}}(\mathbf{y}|\mathbf{x},\mathbf{v})$.
However, for conciseness, the conditioning variables are omitted in subsequent notations.

\begin{remark}[Parameterized $\pi_\mathit{ref}$]
\label{rm:ref-param}
The reference response distribution $\pi_\mathit{ref}$ could be approximated as a posterior distribution by minimizing the expected negative log-likelihood as in \cref{eq:nvp_nll}.
\begin{equation}
\label{eq:nvp_nll}
    \min \mathbb E_{\mathbf y\sim\pi_\mathit{ref}(\mathbf y)} \left[ -\log \pi_\mathit{ref}(\mathbf y)\right].
\end{equation}
\end{remark}

Given the optimized reference distribution $\pi_\mathit{ref}$, as derived in \Cref{rm:ref-param}, there exists a bijective mapping between the set of non-hallucinated sequences generated by the model and the reference response distribution $\pi_\mathit{ref}$. 
Let $\hat{\mathbf y}$ represent the model-generated sequence. 
Building upon the previous derivations, we formally express this relationship as in \cref{eq:nonhalu}.
\begin{equation} \label{eq:nonhalu}
    \hat{\mathbf y} \text{ is non-hallucinated} \iff \mathbf \exists F:F(\hat{\mathbf y}) \sim \pi_\mathit{ref}(\mathbf y),
\end{equation}
where $F$ is the bijective mapping function that relates $\hat{\mathbf y}$ to $\pi_\mathit{ref}(\mathbf y)$, \ie, we can roughly have a transformed response $\mathbf y \sim \pi_\mathit{ref}(\mathbf y)$ from $\hat{\mathbf y}$, which is distributed according to a probability distribution  $q(\hat{\mathbf y})$.
Since $\pi_\mathit{ref}(\mathbf y)$ is the prior distribution in this case, $q(\hat{\mathbf y})$ can be calculated using the change of variables theorem as in \cref{eq:change-of-var}.
\small
\begin{equation}
\label{eq:change-of-var}
\begin{split}
    \log q(\hat{\mathbf y}) &= \log \pi_\mathit{ref}(\mathbf y) + \log \left(\left| \det \left( \dfrac{\partial F(\hat{\mathbf y})}{\partial \hat{\mathbf y}}\right)\right|\right),
\end{split}
\end{equation}\normalsize
where $\det \left( \frac{\partial F(\hat{\mathbf y})}{\partial \hat{\mathbf y}}\right)$ denotes the determinant of the Jacobian matrix at $\hat{\mathbf y}$. 

To learn the mapping function $F$, we propose to employ a simple yet effective flow-based generative model. The model is parameterized by $\theta_F$, and is learned by minimizing the following negative log-likelihood function as in \cref{eq:nll:flow}.
\begin{equation}
\begin{split}
    \theta_F^* &=\arg\min_{\theta_F} \mathbb E_{\hat{\mathbf y}\sim q(\hat{\mathbf y})} \left[ -\log q(\hat{\mathbf y})\right].
\end{split}
\label{eq:nll:flow}
\end{equation}

\noindent
{\bfseries Bijective Metric.} 
Following the notations above, $\hat{\mathbf y}$ and $\mathbf y$ represent the predicted token sequence of the model and the actual ground-truth sequence, respectively. 
With the optimal parameterized model $\theta_F^*$, the bijective transformation $F$ stated at \cref{eq:change-of-var} can be written as $F(\hat{\mathbf y}; \theta_F^*) \sim \pi_\mathit{ref}(\mathbf y)$. 
Therefore, we propose a bijective metric as the complementary objective loss function for the instruction fine-tuning phase. 
Let this metric be denoted as $\mathcal L_\text{B}$. 
Derived from \cref{eq:nll:flow}, the computation of $\mathcal L_\text{B}$ is formally written as in \cref{eq:bijectequ}.
\small
\begin{equation}
\begin{split}
    \mathcal L_\text{B}&= \mathbb E_{\hat{\mathbf y}\sim q(\hat{\mathbf y})} \left[ -\log q(\hat{\mathbf y})\right] \\
    &= \mathbb E_{\hat{\mathbf y}\sim q(\hat{\mathbf y})} \left[-\log \pi_\mathit{ref}(\mathbf y) - \log \left(\left| \det \left( \dfrac{\partial F(\hat{\mathbf y}; \theta_F^*)}{\partial \hat{\mathbf y}}\right)\right|\right)\right].
\end{split} \label{eq:bijectequ}
\end{equation}
\normalsize

As illustrated in \cref{fig:bijec}, this bijective metric $\mathcal L_\text{B}$ quantifies the degree of hallucination induced by the LVLMs. 
Intuitively, the sequence exhibiting a higher degree of hallucination will be substantially distanced from the reference response distribution $\pi_\mathit{ref}$. 
Conversely, sequences that establish a bijective correspondence with a nearby region of $\pi_\mathit{ref}$ are expected to be less susceptible to hallucination.

The loss function for the fine-tuning procedure would be the unification of the cross-entropy loss computed for the original version of LLaVA v1.5~\cite{liu2024improved} and this additional bijective loss for the bijection-driven decoding process.

Considering this flow-based bijective loss for the decoding stage as complementary, the objective depicted in \cref{eq:llm:nll} can be formally re-written as in \cref{eq:lossfn}.
\small
\begin{equation}
    \theta^* =\arg\min_\theta \left(\mathbb E_{(\mathbf x, \mathbf v, \mathbf y) \in \mathcal D} \left[- \log p_\theta(\mathbf y|\mathbf x, \mathbf v)\right]  + \lambda\cdot \mathcal L_\text{B}\right) ,
    \label{eq:lossfn}
\end{equation}
\normalsize
where $\lambda$ represents the coefficient value. In our experiment, we use $\lambda=10^{-6}$ to achieve the balance across learning objective terms.

\subsection{Instruction Fine-tuning with Bijective Metric}
\label{sec:method:finetune}
This section delves into the formation of the reference response distribution $\pi_\mathit{ref}$ to learn the correspondence between each response and its degree of hallucination. Additionally, this section discusses the application of our proposed bijective metric in instruction fine-tuning.

\noindent
{\bfseries $\pi_\mathit{ref}$ Formation.}
The dataset for training distribution modeled on reference responses is required to have the context length $C$ and the vocab size $V$. 
Consequently, we construct a dataset consisting of multidimensional vectors, each of which is a one-hot encoded vector denoted as $\mathbf x \in \mathbb [0,1]^{1\times C\times V}$. 
Taking the implicit sparsity of the data into consideration during the construction of the instruction-tuning dataset, we use bilinear interpolation to scale down the vector size during our experiments. 
Eventually, we construct a dataset $\mathcal D_\pi = \{\mathbf x_i\}_{i=1}^{M}$, with $\mathbf x \in \mathbb [0,1]^{1\times H\times W}$ ($H$ and $W$ is the down-scaled size) and $M$ is the total number of reference responses in our given data sample. 
The veracity of reference responses is paramount so that they can originate from a specific instruction-fine-tuning dataset for LVLMs.
The rationale behind this is that the ultimate objective of this process is to guide the model to respond accurately and consistently in the context of meticulously crafted conversations.

\noindent
{\bfseries Instruction Fine-tuning.} \cref{fig:framework} depicts the framework of the fine-tuning process with the integration of our flow-based bijection model. 
Specifically, we adopt the same settings as the original baseline~\cite{liu2024improved}, with only minor modifications when incorporating the bijective loss as the objective loss function, as outlined in \cref{eq:lossfn}. 
To attain this stage, the bijection model must be trained beforehand on dataset $\mathcal D_\pi$ as outlined above.
During the instruction fine-tuning phase, the parameters of this model are frozen to maintain the correlation between each model-generated response and its level of hallucination.
Furthermore, by the baseline, we only train the parameters of the large language model within the framework.
Overall, given an input image and a text query, the LVLM must generate a response that is then compared with the reference response distribution $\pi_\mathit{ref}$.
This distribution guides the response to adhere to the appropriate pattern compared to the learned sequences.
By curbing the behavior of the decoding process in this manner, the response will be less prone to hallucinations, thereby improving the model’s performance.

\section{Experimental Results}
\label{sec:exp}

\subsection{Benchmarks and Metrics}

{\bfseries POPE}~\cite{li-etal-2023-evaluating-pope} The Polling-based Object Probing Evaluation (POPE) is a benchmark tailored for assessing object hallucination issues in LVLMs. Specifically, the models are inquired to arrive at an answer if a determined object exists in a given image. To query the models this way, the question format is \texttt{"Is there a <object> in the image?"}. Existent and non-existent objects have a balanced ratio within this benchmark. The evaluation consists of three distinct settings based on the construction of negative samples: random, popular, and adversarial. In random split, the selection of objects was randomly made from the entire dataset. In the popular setting, high-frequency objects are selected to be assessed. In the adversarial split, co-occurring objects are collected to evaluate the LVLMs.

\noindent
{\bfseries CHAIR}~\cite{rohrbach-etal-2018-object-chair} The Caption Hallucination Assessment with Image Relevance (CHAIR) is a rule-based metric to evaluate object hallucination for caption generation tasks. Particularly, CHAIR calculates the proportion of objects mentioned in the image's description that are not acquainted with the ground-truth labels. CHAIR benchmark contains two discrete setups, including $\operatorname{CHAIR}_S$ that probes the sentence-level hallucinated objects and $\operatorname{CHAIR}_I$ that gauges on image level. These two settings can be codified as the following:
\small
\begin{gather}
    \operatorname{CHAIR}_S =\dfrac{|\{\text{captions with hallucinated objects}\}|}{|\{\text{all captions}\}|}, \\
    \operatorname{CHAIR}_I =\dfrac{|\{\text{hallucinated objects}\}|}{|\{\text{all mentioned objects}\}|}.
\end{gather}
\normalsize

\subsection{Dataset and Baseline Models}
\label{sec:exp:dataset}
\noindent
{\bfseries Dataset.} We use instruction fine-tuning data\footnote{https://huggingface.co/datasets/liuhaotian/LLaVA-Instruct-150K} published by LLaVA v1.5~\cite{liu2024improved} for our empirical experimentation. In particular, we use \texttt{llava\_v1\_5\_mix665k} version to create the dataset $\mathcal D_\pi$ for $\pi_\mathit{ref}$ formation.
It consists of over 665,000 comprehensive conversations between a user and an intelligent assistant. 
For fair comparisons, we use the same dataset version for our instruction fine-tuning step.

\begin{figure*}[!t]
    \begin{subfigure}{0.5\textwidth}
        \includegraphics[width=\linewidth]{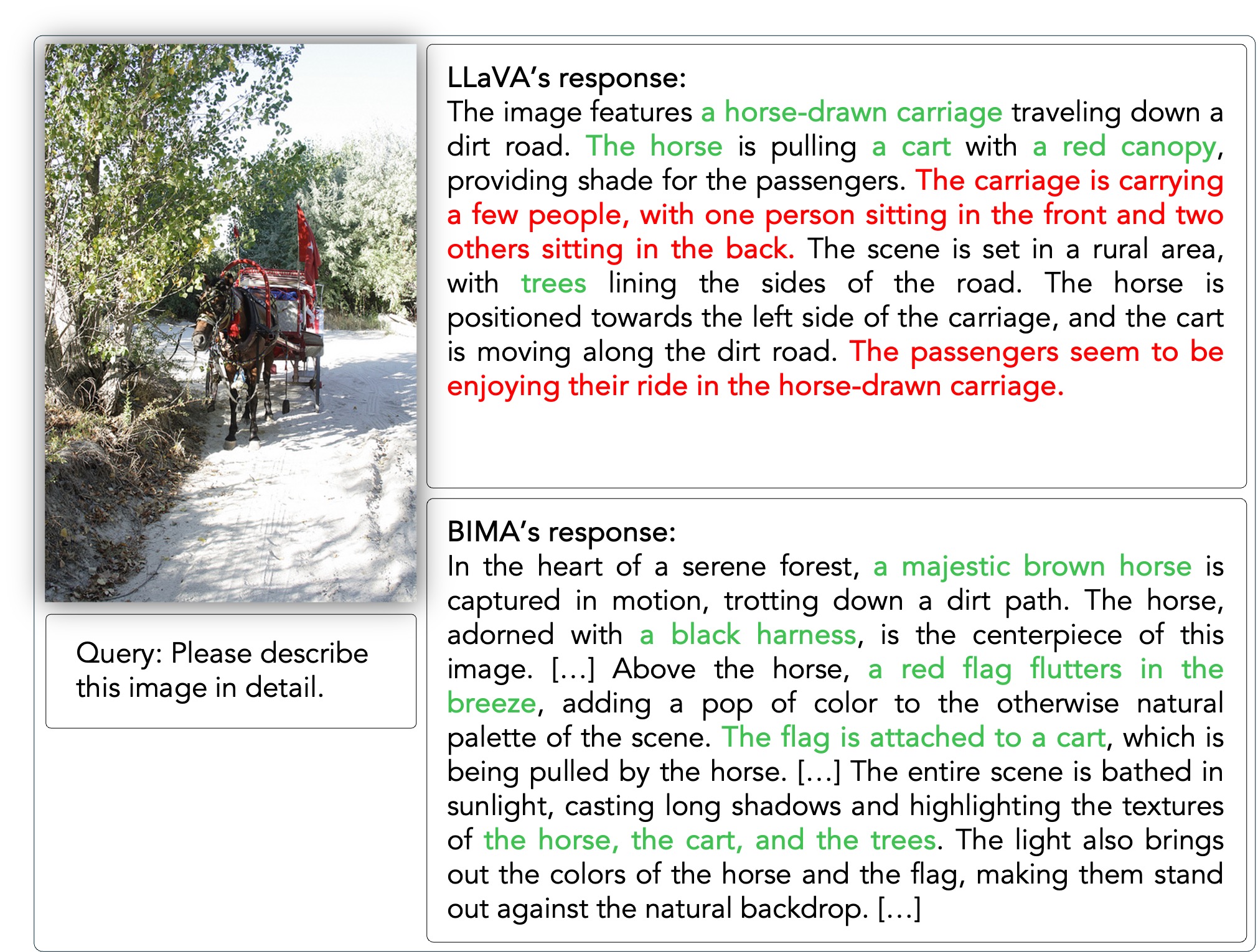} 
        \caption{}
        \label{fig:quali:1}
    \end{subfigure}
    \begin{subfigure}{0.5\textwidth}
        \includegraphics[width=\linewidth]{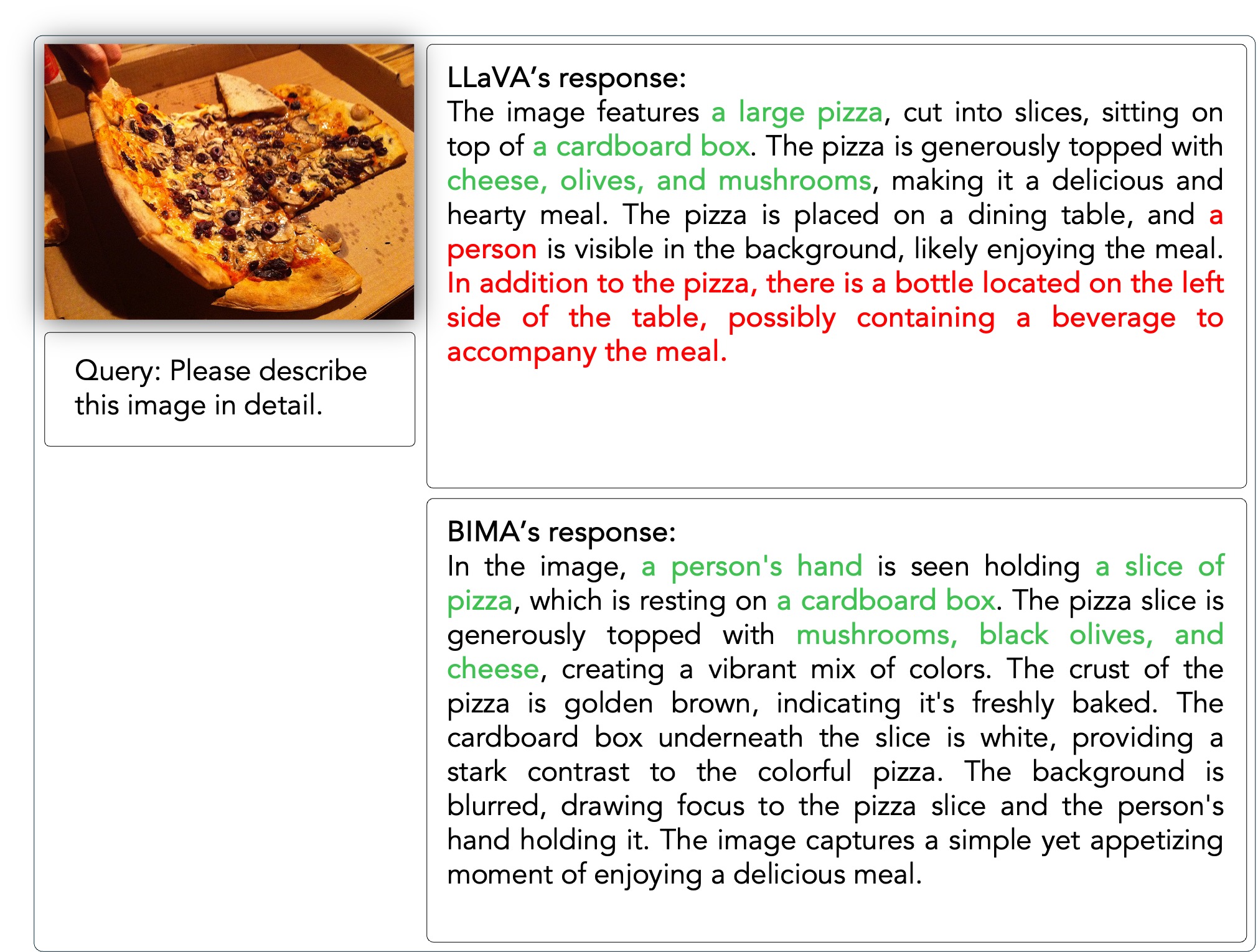}
        \caption{}
        \label{fig:quali:2}
    \end{subfigure}
    
    \caption{Qualitative results of our experimentation to assess the performance of \itsname approach in image caption generation task. We compare pairs of responses received from LLaVA v1.5~\cite{liu2024improved} and our proposed method.}
    \label{fig:quali}
    \vspace{-4mm}
\end{figure*}

\noindent
{\bfseries Baseline LVLMs.}
LLaVA v1.5~\cite{liu2024improved}, an open-source and robust large vision-language model, is widely utilized in research investigating hallucinations. 
This model integrates a visual encoder with a large language model through a projection constructed using either a linear or multi-layer perceptron network. 
By projecting visual tokens into textual representation space, the model effectively learns the correlation between the two modalities.
Building upon the remarkable visual instruction tuning paradigm introduced in ~\cite{liu2023llava}, LLaVA v1.5 devised an AnyRes technique to scale the model with higher-resolution images. Given the computational resource constraints, we opted to utilize LLaVA v1.5 7B in our study.

\subsection{Implementation}
Our experiments adopted the framework from LLaVA v1.5~\cite{liu2024improved} as the baseline LVLM. We used CLIP-ViT-L-14 (336px) for the vision encoder and Vicuna-7B v1.5~\cite{vicuna2023} for the large language model in the framework.
Additionally, we used the two-layer perceptron for the projection between visual and textual representation space. 
Besides, we used the RealNVP~\cite{dinh2017realnvp} model to train the flow-based bijective module, with the number of scales set to 4 and the number of blocks set to 8. 
For training the bijective transformation function, we used the dataset with $C=2,048$ and $V=32,002$. 
Furthermore, the vectors trained in this step were resized into the target shape of $128 \times 128$. 
For instruction fine-tuning, we trained the entire dataset mentioned at \cref{sec:exp:dataset} with one epoch.
The models in our experiments were trained using 32 A100 GPUs with 40GB of VRAM each.
For the image captioning evaluation benchmark (CHAIR), we set the maximum number of new tokens to 512 for all the methods.

\subsection{Quantitative Results}

\begin{table}[!b]
    \small
    \centering
    \setlength{\tabcolsep}{5mm}{
    \begin{tabular}{@{}l*{2}{c}}
        \toprule
        \multicolumn{1}{c}{\bfseries Method}
        & $\operatorname{CHAIR}_S \downarrow$ & $\operatorname{CHAIR}_I \downarrow$ 
        \\
        \midrule
        Greedy & 45.0 & 14.7 \\ 
        Nucleus & 48.8 & 14.2 \\ 
        Beam Search & 48.8 & 13.9 \\ 
        {OPERA}~\cite{huang2024opera} &  44.6 &  12.8 \\ 
        ICD~\cite{icd} &  47.4 &  13.9 \\ 
        VCD~\cite{leng2024vcd} &  46.8 &  13.2 \\ 
        SID~\cite{huo2025sid} &  44.2 &  12.2  \\ 
        ProjectAway~\cite{jiang2025interpreting} &\underline{42.0}	&\underline{12.2}\\ 
        \itsname (ours) &  \textbf{34.4}&  \textbf{9.6}\\
        \bottomrule
    \end{tabular}
    }
    \caption{Evaluation results of our proposed \itsname compared to prior decoding-based approaches on the CHAIR benchmark~\cite{rohrbach-etal-2018-object-chair}. The experiments use LLaVA v1.5~\cite{liu2024improved} as the baseline. The reported metrics compromise $\operatorname{CHAIR}_S$ and $\operatorname{CHAIR}_I$. Best values are in \textbf{bold} and \underline{underlined}.}
    \label{tab:exp:chair512}
\end{table}

\begin{table*}[!t]
    \small
    \centering
    \setlength{\tabcolsep}{3mm}{
    \begin{tabular}{@{}l*{6}{c}}
        \toprule
        \multicolumn{1}{c}{\bfseries Setting}
        & \multicolumn{2}{c}{Random} 
        & \multicolumn{2}{c}{Popular} 
        & \multicolumn{2}{c}{Adversarial} 
        \\ 
        \cmidrule(r){1-1}
        \cmidrule(rl){2-3}
        \cmidrule(rl){4-5}
        \cmidrule(l){6-7}
        \multicolumn{1}{c}{\bfseries Method} & Accuracy $\uparrow$ & F1 Score $\uparrow$ & Accuracy $\uparrow$ & F1 Score $\uparrow$ & Accuracy $\uparrow$ & F1 Score $\uparrow$
        \\
        \midrule
        Nucleus &84.77	&82.28	&79.98	&79.34	&76.03	&76.26 \\ 
        Beam Search & 88.81 &88.52 & 82.76 & 83.36 &79.11 &80.92 \\ 
        OPERA~\cite{huang2024opera} &\underline{88.85}	&\underline{88.67}	&82.77	&83.4	&79.16	&80.93 \\ 
        ICD~\cite{icd} &87.97	&87.84	&84.03	&84.22	&80.21	&80.97 \\ 
        VCD~\cite{leng2024vcd} & 87.02	&86.96	&83.53	&84.56	&78.12	&80.16 \\ 
        SID~\cite{huo2025sid} &\textbf{89.46}	&\textbf{89.62}	&\underline{85.13}	&\textbf{85.94}	&\textbf{83.24}	&\underline{82.21} \\ 
        \itsname (ours) & 88.03	&87.11	&\textbf{85.73}	&\underline{85.00}	&\underline{83.13}	&\textbf{83.07} \\
        \bottomrule
    \end{tabular}
    }
    \caption{Evaluation results of our proposed \itsname in comparison with various decoding-based approaches on the POPE MSCOCO benchmark~\cite{li-etal-2023-evaluating-pope}, including three splits: random, popular, and adversarial. The baseline LLaVA v1.5~\cite{liu2024improved} is employed in the experiment. The reported metrics encompass {\em Accuracy} and {\em F1 Score}. Best values are in \textbf{bold} and \underline{underlined}. 
    }
  
  \label{tab:exp:pope}
  \vspace{-4mm}
\end{table*}

In this section, we evaluate the performance of our flow-based decoding as a hallucination mitigator while integrated with LVLMs on long descriptions and simplified visual question-answering benchmarks.

\noindent
{\bfseries Effectiveness on CHAIR benchmark.}
As illustrated in \cref{tab:exp:chair512}, our method shows a remarkable reduction in hallucination by both scores of $\operatorname{CHAIR}_S$ and $\operatorname{CHAIR}_I$. 
Particularly, under the same setting of maximum new generated tokens (which is 512), our model achieves $\operatorname{CHAIR}_S$ of $34.4\%$ and $\operatorname{CHAIR}_I$ of $9.6\%$. 
Our method surpasses prior approaches by a significant margin, with $7.6\%$ and $2.6\%$ improvements in $\operatorname{CHAIR}_S$ and $\operatorname{CHAIR}_I$, respectively.
These results demonstrate the effectiveness of our proposed \itsname when employed as a hallucination mitigator in the image caption generation task.
Due to the inherent nature of long-form captions, our method effectively learns to produce non-hallucinated captions via bijective correspondence initialized by the bijection model.

\noindent
{\bfseries Effectiveness on POPE benchmark.}
Through our experiments, our proposed \itsname approach achieves competitive scores on the POPE benchmark.
As portrayed in \cref{tab:exp:pope}, our method outperforms most of the prior approaches with regard to popular and adversarial splits. 
Specifically, compared to SID~\cite{huo2025sid}, our method reaches an accuracy of $85.73\%$ ($+0.60\%$) on the popular setting and an F1-score of 83.07\% ($+0.86\%$) on the adversarial setting, while maintaining the compelling results on popular split's F1 score and adversarial setting's accuracy (with $85.00\%$ and $83.13\%$ respectively). 
Although the POPE benchmark is focused on visual question answering, which implies that model-generated responses have limited length (the number of maximum new tokens is below 30), our proposed \itsname method still demonstrates its efficacy.

\noindent
{\bfseries Discussion.}
Based on the empirical results conducted on different benchmarks, our proposed \itsname approach appears to exhibit superior performance in the context of longer responses. 
This is because extended responses provide more context to the bijection model when mapping the response itself to its corresponding level of hallucination. 
To further enhance the capability of alleviating hallucinated outputs, our proposed \itsname could be integrated with complementary techniques, such as considering the significance of visual tokens during decoding~\cite{huo2025sid}.

\subsection{Qualitative Results}

We primarily analyze the performance of our proposed \itsname approach in reducing hallucination in the image caption generation task.
\cref{fig:quali} illustrates some of our qualitative results on COCO~\cite{lin2014microsoft}.
Overall, \itsname shows its superiority in the image captioning task since it generates longer but more robust against hallucinations. 
In \cref{fig:quali:1}, the baseline model's response exhibits notable deficiencies, particularly in the generation of hallucinated objects that are absent from the scene, such as ``people'' or ``passengers.'' 
In contrast, our model produces more contextually precise captions. 
Specifically, \itsname's response not only retains the core set of primary objects, including ``horse,'' ``cart'' or ``trees,'' but also provides a more comprehensive depiction of the image. This includes recognizing additional objects (\eg ``a black harness'') or capturing finer visual attributes, such as the scene's illumination (\eg ``...is bathed in sunlight'').
Similarly, as depicted in \cref{fig:quali:2}, while the baseline model generates captions that accurately identify primary objects such as ``pizza,'' ``cardboard box'' and relevant toppings like ``cheese,'' ``olives'' and ``mushrooms,'' it also inaccurately hallucinates nonexistent elements in the scene, such as ``a person'' or ``a bottle.'' 
Conversely, \itsname's responses demonstrate greater robustness by reliably identifying solely objects that are truly present in the image. 
Moreover, it provides a more precise and context-aware description (\eg ``a person's hand''), consolidating its capability to mitigate hallucinations and produce responses that are more aligned with the visual content.
These findings further validate \itsname's efficacy in generating high-quality, contextually accurate captions, solidifying its position as a robust approach for vision-language understanding tasks.

\section{Conclusion and Limitation}
\label{sec:concl}
This work has endeavored to mitigate the hallucination induced by large vision-language models. 
We proposed {\em\itsname}, a compelling decoding approach using a new bijective maximum likelihood learning. 
It exploits the normalizing flow theory and establishes the interrelationship between the model-generated responses and the ground-truth distribution, thereby defining the degree of hallucination in the responses.
We also introduced an instruction fine-tuning method with the integration of this novel bijective metric to mitigate hallucinations in LVLMs.
With the empirical results on POPE and CHAIR benchmarks, our proposed \itsname shows its effectiveness in reducing hallucinations and improving LVLMs' reliability.

\noindent
{\bfseries Limitations.} 
While this study employs a bijection-based approach to improve the performance of LLaVA v1.5, a more comprehensive evaluation across various LVLMs is needed to assess the generalizability of this work. 
In addition, incorporating the larger instruction-tuning dataset will provide a more representative analysis of the reference response distribution, offering deeper insights into the model's behavior. 
Future work can focus on exploring these directions to further evaluate the effectiveness of our approach in mitigating hallucinations in LVLMs.

\noindent
{\bfseries Acknowledgment.}
This material is based upon work supported by the National Science Foundation under Award No. OIA-1946391. We also acknowledge the Arkansas High-Performance Computing Center for providing GPUs.

{
    \small
    \bibliographystyle{ieeenat_fullname}
    \bibliography{main}
}


\end{document}